\documentclass[11pt,a4paper]{article}
\usepackage[hyperref]{ranlp2025}
\usepackage{times}
\usepackage{latexsym}

\usepackage{microtype}
\usepackage{booktabs}
\usepackage{graphicx}
\usepackage{caption}
\usepackage{subcaption}
\usepackage{float}
\usepackage{makecell}

\aclfinalcopy % Uncomment this line for the final submission
\usepackage{multirow}
\usepackage{csvsimple}

\usepackage{ifthen}
\newcommand*{\GottBERT}[2]{\ifthenelse{\equal{#2}{last}}{$\mathrm{GottBERT}_{#1}^{\crosssymbol}$}{$\mathrm{GottBERT}_{#1}$}}
\newcommand*{\GottBERTf}[2]{\ifthenelse{\equal{#2}{last}}{$^{\mathrm{f}}\mathrm{GottBERT}_{#1}^{\mathrm{\crosssymbol}}$}{$^{\mathrm{f}}\mathrm{GottBERT}_{#1}$}}

\newcommand{\GottBERTt}{$\mathrm{GottBERT}$}

\title{PortBERT: Navigating the Depths of Portuguese Language Models}

\author{
 \textbf{Raphael Scheible-Schmitt\textsuperscript{1,2,3}},
 \textbf{Henry He\textsuperscript{1}},
 \textbf{Armando B. Mendes \textsuperscript{2}}
\\
\\
 \textsuperscript{1}School of Computation, Information and Technology, Technical University of Munich, Munich, Germany\\
 \textsuperscript{2}IS\textsuperscript{2}E - Intelligent Systems, Science and Engineering, LIACC polo on Azores University, Ponta Delgada, Portugal\\
 \textsuperscript{3}Institute of General Practice, Faculty of Medicine and Medical Center, University of Freiburg
\\
 \small{
   \textbf{Correspondence:} \href{mailto:raphael.scheible@tum.de}{raphael.scheible@tum.de}
 }
}

\date{}

\begin{document}
\maketitle
\thispagestyle{plain}
\begin{abstract}
Transformer models dominate modern NLP, but efficient, language-specific models remain scarce. In Portuguese, most focus on scale or accuracy, often neglecting training and deployment efficiency.
In the present work, we introduce PortBERT, a family of RoBERTa-based language models for Portuguese, designed to balance performance and efficiency. Trained from scratch on over 450 GB of deduplicated and filtered mC4 and OSCAR23 from CulturaX using fairseq, PortBERT leverages byte-level BPE tokenization and stable pre-training routines across both GPU and TPU processors. We release two variants, PortBERT\textsubscript{base} and PortBERT\textsubscript{large}, and evaluate them on ExtraGLUE, a suite of translated GLUE and SuperGLUE tasks.  Both models perform competitively, matching or surpassing existing monolingual and multilingual models. Beyond accuracy, we report training and inference times as well as fine-tuning throughput, providing practical insights into model efficiency. PortBERT thus complements prior work by addressing the underexplored dimension of compute-performance tradeoffs in Portuguese NLP. We release all models on Huggingface and provide fairseq checkpoints to support further research and applications.
\end{abstract}

\section{Introduction}

The development of neural language models has profoundly shaped natural language processing (NLP), particularly through the advent of transformer-based architectures such as BERT~\citep{devlin_bert_2019} and its optimized variant RoBERTa~\citep{liu_roberta_2019}. These models, which learn contextualized word representations via self-supervised pretraining, have become foundational across a wide range of NLP tasks. While early efforts prioritized English or multilingual solutions, research has shown that language-specific pretraining on high-quality, monolingual corpora often yields superior results for the target language~\citep{delobelle_robbert_2020,scheible-etal-2024-gottbert}.

In Portuguese NLP, monolingual transformer models such as BERTimbau~\citep{souza2020bertimbau} and AlBERTina~\citep{Rodrigues_2023} have marked important milestones. More recently, multilingual alternatives like XLM-RoBERTa~\citep{chan_xlm-roberta_2020} and EuroBERT~\citep{boizard2025eurobertscalingmultilingualencoders} have demonstrated strong cross-lingual performance by scaling up to billions of parameters. EuroBERT, in particular, follows the "Modern BERT" framework~\citep{warner2024smarterbetterfasterlonger}, which revisits encoder-based models with streamlined design and improved training efficiency. While decoder-only models continue to dominate general-purpose NLP, these developments show that encoder-based masked language models (MLMs) remain competitive and relevant.

However, many of these advancements come at considerable computational cost. As NLP systems move closer to real-world applications, ranging from chatbots and document pipelines to tasks such as named entity recognition, sentence classification, or part-of-speech tagging, efficiency becomes a central concern. Models deployed in production must often meet strict requirements in terms of latency, memory usage, and energy consumption. Prior work has shown that compact transformer models can offer significant speed-ups with minimal impact on performance~\citep{sanh2020distilbertdistilledversionbert, jiao-etal-2020-tinybert}. Yet, most Portuguese models focus primarily on accuracy, offering limited insight into training efficiency, hardware utilization, or deployment tradeoffs.

To address this gap, we introduce PortBERT, a family of RoBERTa-based encoder models tailored for Portuguese. PortBERT is trained from scratch on over 450 GB of deduplicated text from CulturaX~\citep{nguyen2023culturaxcleanedenormousmultilingual}, combining data from mC4~\citep{xue-etal-2021-mt5} and OSCAR23~\citep{jansen2022perplexedqualityperplexitybasedmethod}. Following \citet{scheible-etal-2024-gottbert}, we construct a byte-level BPE vocabulary with 52k tokens using Hugging Face’s tokenizer tools, which helps improve token efficiency and compression, an effect observed in prior work on Dutch and German~\citep{delobelle_robbert_2020,scheible-etal-2024-gottbert}.

Pretraining is performed using the fairseq framework: the base variant is trained on 8 NVIDIA A40 GPUs, and the large variant on a TPUv4-128 pod. PortBERT retains the standard RoBERTa architecture without architectural modifications like sparse attention or extended context. Instead, it emphasizes a balanced design that prioritizes pretraining efficiency, inference throughput, and downstream accuracy. While not designed to match the scale of models like EuroBERT~\citep{boizard2025eurobertscalingmultilingualencoders} or decoder-based LLMs, PortBERT offers a robust, reproducible, and accessible alternative for practical Portuguese NLP.

The main contributions of this study are:
\begin{itemize}
    \item We provide two variants, PortBERT\textsubscript{base} and PortBERT\textsubscript{large}, trained respectively on GPUs and a TPUv4 pod, and release both models under an open-source license.

    \item We evaluate PortBERT on the ExtraGLUE benchmark, showing that both models perform competitively.

    \item We report training time for both pretraining and downstream fine-tuning, and include throughput metrics for fine-tuning to support transparent evaluation of efficiency.
\end{itemize}

% - NLP in general, contextual blabla
% - Portugese nlp

% - https://link.springer.com/chapter/10.1007/978-981-96-1551-3_20
% - https://link.springer.com/chapter/10.1007/978-3-031-21753-1_28
% - https://aclanthology.org/2024.propor-1.38/
% - Albertina: https://arxiv.org/abs/2305.06721
% - modern bert: https://arxiv.org/abs/2412.13663
% - EuroBERT: https://huggingface.co/blog/EuroBERT/release
% - GottBERT: https://aclanthology.org/2024.emnlp-main.1183/
% - GeistBERT: https://arxiv.org/abs/2506.11903
 
\section{Related works}

In recent years, a growing number of transformer-based language models have been developed for Portuguese. These include both monolingual models trained specifically on Portuguese corpora and multilingual models that support a wide range of languages. Table~\ref{tab:related} summarizes these models, their architectures, and training data sources.

BERTimbau~\citep{souza2020bertimbau} was one of the first monolingual BERT-style models for Portuguese, available in base and large versions. It was trained on a mix of BrWaC~\cite{wagner2018brwac}, Portuguese Wikipedia, and a news corpus using whole-word masking (WWM) over one million steps.

AiBERTa\footnote{\url{https://huggingface.co/AiBERTa/aiberta-d-2000M-random}}~\citep{10.1007/978-3-031-21753-1_28, 10.1007/978-981-96-1551-3_20} follows a RoBERTa-style architecture and is trained on a curated subset of Portuguese periodical websites archived in \texttt{Arquivo.pt}, a national web archive. These periodicals range from national newspapers like Público to smaller regional outlets, providing well-written and structurally consistent Portuguese text.

AlBERTina~\citep{Rodrigues_2023} adopts the ALBERT architecture~\citep{lan2020albertlitebertselfsupervised}, introducing parameter-sharing and embedding factorization. The models were trained on the January 2023 version of OSCAR, as well as DCEP, Europarl, and ParlamentoPT. Separate variants exist for Brazilian and European Portuguese.

RoBERTa PT~\citep{roberta_pt} was trained on 10 million English and 10 million Portuguese sentences from the OSCAR corpus. Despite its bilingual setup and relatively small training corpus, the model is widely cited and has been evaluated in various Portuguese NLP tasks.

RoBERTaCrawlPT and RoBERTaLexPT~\citep{garcia-etal-2024-robertalexpt} are both RoBERTa-based models developed for Portuguese. RoBERTaCrawlPT uses CrawlPT, a combined corpus comprising BrWaC, CC100-PT, and OSCAR23-PT. RoBERTaLexPT targets legal-domain applications and adds LegalPT, a corpus aggregating diverse legal documents totaling up to 125~GiB.
% do you think is relevant to mention amalia, the newest LLM for european pt, chatamalia.ia ?

Among multilingual models, XLM-RoBERTa~\citep{chan_xlm-roberta_2020} can be used for Portuguese tasks. It is trained on 2.5~TB of filtered Common Crawl data in over 100 languages, including Portuguese.

EuroBERT~\citep{boizard2025eurobertscalingmultilingualencoders} is a more recent multilingual encoder model that spans 15 European languages, including Portuguese. It follows the Modern BERT architecture~\citep{warner2024smarterbetterfasterlonger}, with design choices optimized for scalability and efficiency. Its training data includes CulturaX~\citep{nguyen2023culturaxcleanedenormousmultilingual}, FineWeb~\citep{penedo2024finewebdatasetsdecantingweb}, EuroLLM~\citep{martins2024eurollmmultilinguallanguagemodels}, and code-related corpora such as The Stack v2~\citep{lozhkov2024starcoder2stackv2} and Proof-Pile-2~\citep{azerbayev2024llemma}.

While many Portuguese models report strong downstream performance, few document training efficiency or hardware usage. PortBERT complements this work by offering initial insights into these often underreported aspects.

% \begin{table*}[h!tbp]
% \centering
% \begin{tabular}{lcccc}
% \hline
% \bf Model & \bf Architecture & \bf \#Lang &  \bf Corpus Source \\
% \hline
% BERTimbau\textsubscript{base}         & BERT     & 1  & Wikipedia + OSCAR \\
% BERTimbau\textsubscript{large}        & BERT     & 1  & Wikipedia + OSCAR \\
% AlBERTina 100M PTPT                   & ALBERT  & 1  & OSCAR, DCEP, Europarl, and ParlamentoPT        \\
% AlBERTina 100M PTBR                   & ALBERT  & 1  & OSCAR, DCEP, Europarl, and ParlamentoPT        \\
% AiBERTa                                & RoBERTa  & 1  & OSCAR + Sabiá + Corpus NILC \\
% RoBERTa PT                             & RoBERTa  & 2 & 10M PT + 10M EN sentences from OSCAR \\
% RoBERTaCrawlPT\textsubscript{base}     & RoBERTa  & 1  & OSCAR + CommonCrawl \\
% RoBERTaLexPT\textsubscript{base}       & RoBERTa  & 1  & OSCAR + legal corpora \\
% EuroBERT 210M                         & Moden BERT  & 15  & OSCAR (filtered, 15 Euro langs) \\
% EuroBERT 610M                         & Modern BERT  & 15  & OSCAR (filtered, 15 Euro langs) \\
% XLM-RoBERTa\textsubscript{base}       & RoBERTa  & 100+ & CommonCrawl, Wikipedia \\
% XLM-RoBERTa\textsubscript{large}      & RoBERTa  & 100+ & CommonCrawl, Wikipedia \\
% \hline
% \end{tabular}
% \caption{\label{tab:related} Overview of Portuguese and multilingual language models. \#Lang indicates the number of languages. Corpus size and sources are based on published metadata.}
% \end{table*}
\begin{table*}[htb]
\centering
\begin{tabular}{lccc}
\toprule
\textbf{Model} & \textbf{Architecture} & \textbf{Language(s)} & \textbf{Training Data Sources} \\
\midrule
BERTimbau & BERT & 1 & \makecell[l]{BrWaC, Wikipedia, news corpora} \\
AiBERTa & RoBERTa & 1 & \makecell[l]{Arquivo.pt (Portuguese periodicals)} \\
AlBERTina PTPT/PTBR & ALBERT & 1 & \makecell[l]{OSCAR 23, DCEP, Europarl, ParlamentoPT} \\
RoBERTa PT & RoBERTa & 2 & \makecell[l]{OSCAR (10M sentences each language)} \\
RoBERTaCrawlPT\textsubscript{base} & RoBERTa & 1 & \makecell[l]{CrawlPT (brWaC, CC100-PT, OSCAR23-PT)} \\
RoBERTaLexPT\textsubscript{base} & RoBERTa & 1 & \makecell[l]{CrawlPT, LegalPT (aggregated legal corpus)} \\
XLM-RoBERTa & RoBERTa & 100+ & \makecell[l]{CommonCrawl (2.5TB, filtered)} \\
EuroBERT & Modern BERT & 15 & \makecell[l]{CulturaX, FineWeb, EuroLLM, The Stack v2,\\ Proof-Pile-2} \\
\bottomrule
\end{tabular}
\caption{Overview of transformer-based language models relevant to Portuguese. The table lists architecture type, language coverage, and training data sources.}
\label{tab:related}
\end{table*}

\section{Methods}

\subsection{Corpus}
% R1 Q3: It would be good to see more detail on the deduplication process by CulturaX? How does it compare to more aggressive strategies like MinHash or SimHash?
To pre-train PortBERT, we used the Portuguese portions of mC4 and OSCAR23 \cite{jansen2022perplexedqualityperplexitybasedmethod}, two large-scale web corpora. The original size of Portuguese mC4 was approximately 453.1 GB, and OSCAR23 contributed 96.9 GB, totaling 550 GB of raw data. To reduce redundancy and improve quality, we relied on the deduplicated and filtered versions provided by CulturaX~\cite{nguyen2023culturaxcleanedenormousmultilingual}, which together amount to 456.6 GB, a size reduction of roughly 17\% (93.4 GB). This large and diverse dataset ensures broad linguistic coverage with reduced duplication and noise compared to raw crawled corpora. CulturaX applied language identification, quality filtering, and deduplication to produce these cleaned subsets.

% To pre-train PortBERT, we compiled a large-scale Portuguese corpus consisting of the most recent versions of mC4 and OSCAR corpora. Specifically, we used the Portuguese portion of the mC4 dataset provided through CulturaX \cite{nguyen2023culturaxcleanedenormousmultilingual}, amounting to approximately 453 GB of compressed text, and the Portuguese subset of OSCAR23 \cite{jansen2022perplexedqualityperplexitybasedmethod}, contributing an additional 96.9 GB. After deduplication, the combined corpus (mC4 + OSCAR) totaled 456.6 GB. This large and diverse dataset was intended to maximize coverage across linguistic domains while minimizing redundancy.
% so OSCAR + mC$ did't grow much from the original OSCAR. Can you compare deduplicated OSCAR with final corpus? Otherwise seems that almost everything are duplicated between both corpus

\subsection{Pre-processing}
RoBERTa employs the byte pair encoding (BPE) tokenizer originally introduced with GPT-2 \cite{radford_language_2019}, which processes raw text directly without requiring pre-tokenization or language-specific tools like Moses \cite{koehn_moses_2007}. While this tokenizer was trained on English corpora, we followed the approach taken for GottBERT \cite{scheible-etal-2024-gottbert} by training a dedicated Portuguese tokenizer. Using 40 GB of randomly sampled Portuguese corpus data, we created a 52k-token vocabulary optimized for the language. Although we did not explicitly measure the impact on file size or task performance for PortBERT, similar adaptations in Dutch~\cite{delobelle_robbert_2020} and German~\cite{scheible-etal-2024-gottbert} have demonstrated benefits in both respects. In our experience, a 40 GB sample is sufficient for the subword distribution to converge, and extending vocabulary training to the full corpus would add considerable overhead with little expected benefit.
% R2Q1: Although the paper computes BPE on a 40GB sample of the whole corpus, better subwords could be learned if the BPE algorithm is trained on the whole corpus. The authors can utilize better data handling techniques to achieve this.

\subsection{Pre-training}
Similar to GottBERT, we pre-trained the PortBERT\textsubscript{base} and PortBERT\textsubscript{large} models using the Fairseq framework. PortBERT\textsubscript{large} was trained on a 128-core TPUv4 pod \cite{jouppi_tpu_2023}, while PortBERT\textsubscript{base} was trained on a cluster of 8 NVIDIA A40 GPUs, using the same training corpus and identical optimization hyperparameters. Mixed-precision training (fp16) was disabled for the GPU setup and not supported by the TPU implementation used, ensuring that both models were trained entirely in full precision (fp32). This controlled setup enables a direct comparison of hardware-level training efficiency across compute architectures, without numerical precision optimizations acting as confounding factors.
Both models were trained on Portuguese OSCAR data using the RoBERTa architecture. The PortBERT\textsubscript{base} model completed training in approximately 27 days (2{,}331{,}939 seconds), while PortBERT\textsubscript{large} required around 6.2 days (531{,}807 seconds). We used the standard RoBERTa pretraining schedule with 100k update steps, a batch size of 8k, a 10k-step warmup, and polynomial learning rate decay. The base model used a peak learning rate of 0.0004, and the large model 0.00015. As with GottBERT, we evaluated after each epoch and stored checkpoints throughout training. However, since the dataset size only permitted approximately four epochs, the final checkpoint coincided with the best-performing one.

\subsection{Downstream Tasks}
Based on the pre-trained BERT models, we fine-tuned several downstream tasks using the training scripts provided by Huggingface \cite{wolf_huggingfaces_2020}. Hyperparameter optimization was performed via grid search, focusing on batch size and learning rate. Each task was trained for a maximum of 10 epochs, and the experiments were orchestrated using NNI \cite{nni} on NVIDIA A40 GPUs.

To assess model performance, each downstream task was fine-tuned 28 times using different combinations of batch sizes and learning rates. Since no separate test set was available, we selected the best-performing checkpoint based on validation set scores. The final performance figures reported for each model and task reflect the best result among these 28 validation-based runs. For comparison, we benchmarked our models against eleven other Portuguese language models.

We evaluated the models on ExtraGLUE~\cite{santos_performance_2025}, a Portuguese adaptation of the GLUE benchmark. This suite consists of selected tasks from GLUE and SuperGLUE that were automatically translated into Portuguese, enabling language-specific assessment and ensuring that model performance reflects capabilities in the target language context.

To account for varying input lengths across tasks, we configured the maximum input sequence length individually per task based on the maximum observed input lengths after tokenization across all evaluated models: 192 tokens for MRPC and WNLI, 320 tokens for STS-B, and 512 tokens for RTE. This ensured full coverage of the datasets while avoiding unnecessary padding and memory overhead.

\paragraph{STS-B}
The Semantic Textual Similarity Benchmark (STS-B) task evaluates the model’s ability to assess the semantic similarity between two sentences. Each sentence pair is assigned a similarity score ranging from 0 (completely dissimilar) to 5 (semantically equivalent). Following standard practice, we report the mean of Pearson and Spearman correlation coefficients between predicted and gold scores.

\paragraph{RTE}
The Recognizing Textual Entailment (RTE) task consists of binary classification, where the model must determine whether a given hypothesis logically follows from a provided premise. This task evaluates the model's capacity for inference and semantic reasoning.

\paragraph{WNLI}
The Winograd Natural Language Inference (WNLI) task is a coreference resolution challenge cast as binary entailment. It requires the model to resolve ambiguous pronouns and determine whether a hypothesis follows from a premise. Despite its small size and challenging structure, it is retained for completeness and consistency with GLUE-style benchmarks.

\paragraph{MRPC}
The Microsoft Research Paraphrase Corpus (MRPC) task is a binary classification problem where the model must decide whether two sentences are semantically equivalent. Evaluation is based on both accuracy and F1 score, reflecting the importance of both precision and recall in paraphrase detection.

\subsection{Model Configurations and Properties}
The number of parameters in BERT-like models varies significantly depending on their architecture (see Table \ref{tab:params}). The base version of BERT, such as BERTimbau\textsubscript{base}, has approximately 109 million parameters, while large versions like BERTimbau\textsubscript{large} expand to over 334 million. RoBERTa variants used in Portuguese NLP, such as RoBERTaCrawlPT\textsubscript{base} and RoBERTaLexPT\textsubscript{base}, feature around 125 million parameters, comparable to PortBERT\textsubscript{base} (126M). The large PortBERT model increases this to 357 million, positioning it close to BERTimbau\textsubscript{large} while retaining RoBERTa’s efficiency characteristics.

Multilingual models such as XLM-RoBERTa are designed for cross-lingual tasks, with the base version containing 278 million parameters and the large version 560 million. These parameter counts make them substantially larger than monolingual base models, but beneficial in zero-shot or cross-lingual scenarios \cite{ERONEN2023103250}.

% R1Q5: To will be good to see performance of PortBERT on any zero-shot or cross-lingual benchmarks given the increasing relevance of those settings (depending on the feasibility)?
The AiBERTa and AlBERTina families offer diverse parameter ranges. All AiBERTa variants (regardless of source or domain configuration) have approximately 101 million parameters, with a smaller vocabulary size of 20,000. The AlBERTina models, in contrast, range from 138 million (100M variants) to over 1.5 billion parameters for the 1.5B variants, reflecting a significant increase in capacity and vocabulary size (up to 128,100 tokens). These models serve different use cases depending on the required balance between compute and performance.

Finally, EuroBERT models span from 210 million parameters in the 210M variant to over 2.1 billion in the 2.1B variant. They provide a scalable foundation for multilingual or European-centric tasks, emphasizing both vocabulary coverage and model depth.

\begin{table}[!htbp]
\caption{The size of the vocabulary and the size of the parameters are shown for the model types used in this study. This table does not show other design differences of the models. Values were extracted using Huggingface's transformers library. Models are sorted by number of parameters.}
\label{tab:params}
\centering\small
\begin{tabular}{lcc}
    \hline
    \bfseries Model & \bfseries Vocab Size & \bfseries \#Params \\
    \hline
    \csvreader[late after line=\\]{params.csv}{}%
     {\csvcoli & \csvcolii & \csvcoliii}
    \hline
\end{tabular}
\end{table}

\section{Results}

\subsection{Downstream task evaluation}

Table~\ref{tab:downstream} presents the downstream evaluation results of all Portuguese language models across four ExtraGLUE tasks: STS-B, RTE, WNLI, and MRPC. We report task-specific metrics: Spearman and Pearson correlations for STS-B, accuracy for RTE and WNLI, and both accuracy and F1 for MRPC, alongside the average performance (AVG) across all tasks.

Among the base-sized models, RoBERTaLexPT\textsubscript{base} achieves the highest overall score with an AVG of 80.63, showing strong results particularly in MRPC accuracy (89.46) and F1 (92.34). Close behind is PortBERT\textsubscript{base}, with an AVG of 80.57, outperforming all others in WNLI accuracy (60.56, tied with XLM-R) and ranking second in STS-B with a Spearman score of 87.39 and Pearson of 87.65. BERTimbau\textsubscript{base} shows the best performance in STS-B (88.5 mean), but underperforms slightly in WNLI, holding it back from overall top placement.

RoBERTaCrawlPT\textsubscript{base} and EuroBERT 210m also demonstrate robust overall performance, particularly in RTE and MRPC, with AVG scores above 79.0. Meanwhile, XLM RoBERTa\textsubscript{base} shows competitive results in WNLI (60.56) and MRPC F1 (91.32), though its STS-B score slightly lags behind the top contenders. Legacy models like roBERTa PT perform significantly worse, especially on semantic similarity tasks, confirming the impact of more recent training strategies and data sources.

In the large model category, XLM RoBERTa\textsubscript{large} emerges as the strongest overall model with an AVG of 84.01. It leads all others in STS-B (90.14 mean) and achieves the highest RTE score (82.31), although it underperforms in WNLI. EuroBERT 610m follows closely with an AVG of 83.44, showing outstanding performance in MRPC (94.2 F1, 91.91 accuracy) and the second-best RTE result (78.34).

PortBERT\textsubscript{large} achieves a solid overall score of 82.26, slightly ahead of BERTimbau\textsubscript{large} (82.23). While BERTimbau\textsubscript{large} does not dominate any single task, PortBERT\textsubscript{large} exhibits the highest WNLI accuracy (61.97). BERTimbau\textsubscript{large} stands out with strong STS-B scores (89.5 mean) and competitive MRPC metrics.

Overall, the results validate the effectiveness of the PortBERT models, with both the base and large variants frequently ranking among the top-performing models across tasks. The base model outperforms many existing Portuguese models on average, while the large model achieves results close to the best multilingual transformers. This indicates their robustness and applicability to a range of semantic and inference tasks in Portuguese.

\begin{table*}[htbp]
\begin{center}
\begin{tabular}{lccccccc|c}
    \hline
    \multirow{2}{*}{\bfseries Model} & \multicolumn{3}{c}{\bfseries STS-B (Similarity)} & \bfseries RTE & \bfseries WNLI & \multicolumn{2}{c}{\bfseries MRPC} & \bfseries AVG \\
    & \bfseries Spearman & \bfseries Pearson & \bfseries Mean & \bfseries Acc & \bfseries Acc & \bfseries Acc & \bfseries F1 & \\
    \hline
    \csvreader[late after line=\\]{all_large.csv}{}%
     {\csvcoli & \csvcolii & \csvcoliii & \csvcoliv & \csvcolv & \csvcolvi & \csvcolvii & \csvcolviii & \csvcolix}
    \hline
    \csvreader[late after line=\\]{all_base.csv}{}%
     {\csvcoli & \csvcolii & \csvcoliii & \csvcoliv & \csvcolv & \csvcolvi & \csvcolvii & \csvcolviii & \csvcolix}
    \hline
\end{tabular}
\caption{\label{tab:downstream}
Evaluation results in \%. STSB is reported with Spearman, Pearson, and their mean. RTE and WNLI are classification accuracy. MRPC includes accuracy and F1. The AVG score averages the six metrics: STSB Spearman, STSB Pearson, RTE Acc, WNLI Acc, MRPC Acc, MRPC F1. Bold = best, underlined = second-best per model size. Based on best epoch from 28 runs for max 10 epochs. The AVG score is computed as the unweighted mean across six metrics: STS-B Spearman, STS-B Pearson, RTE accuracy, WNLI accuracy, MRPC accuracy, and MRPC F1.}
\end{center}
\end{table*}
% R1Q4: Can you clarify how performance was averaged in Table 3? Does AVG treat all metrics equally (including F1, accuracy, correlation)?

\subsection{Performance vs. Efficiency}
\label{sec:efficiency}
\begin{figure*}[h!t]
  \includegraphics[width=1\linewidth]{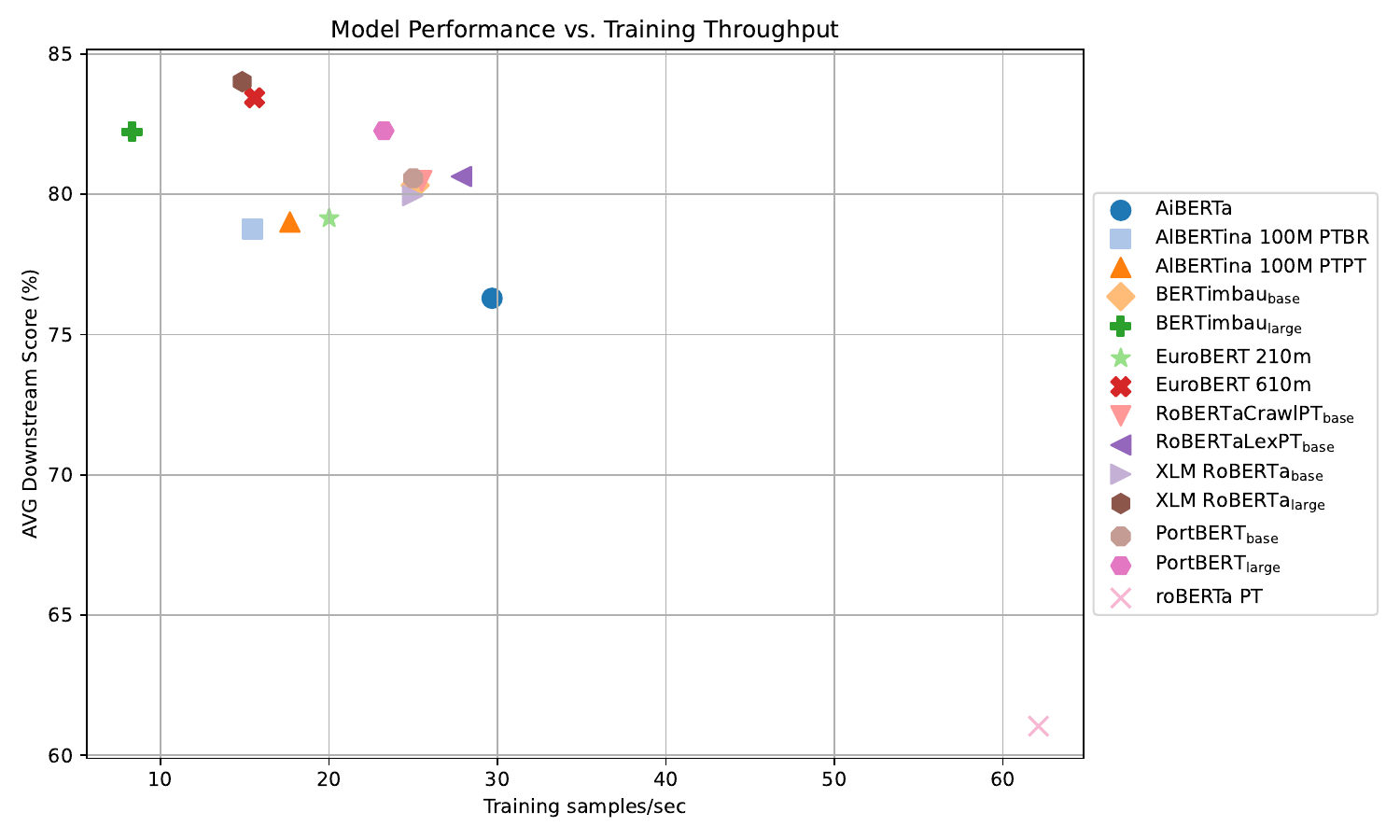} \\
  \includegraphics[width=1\linewidth]{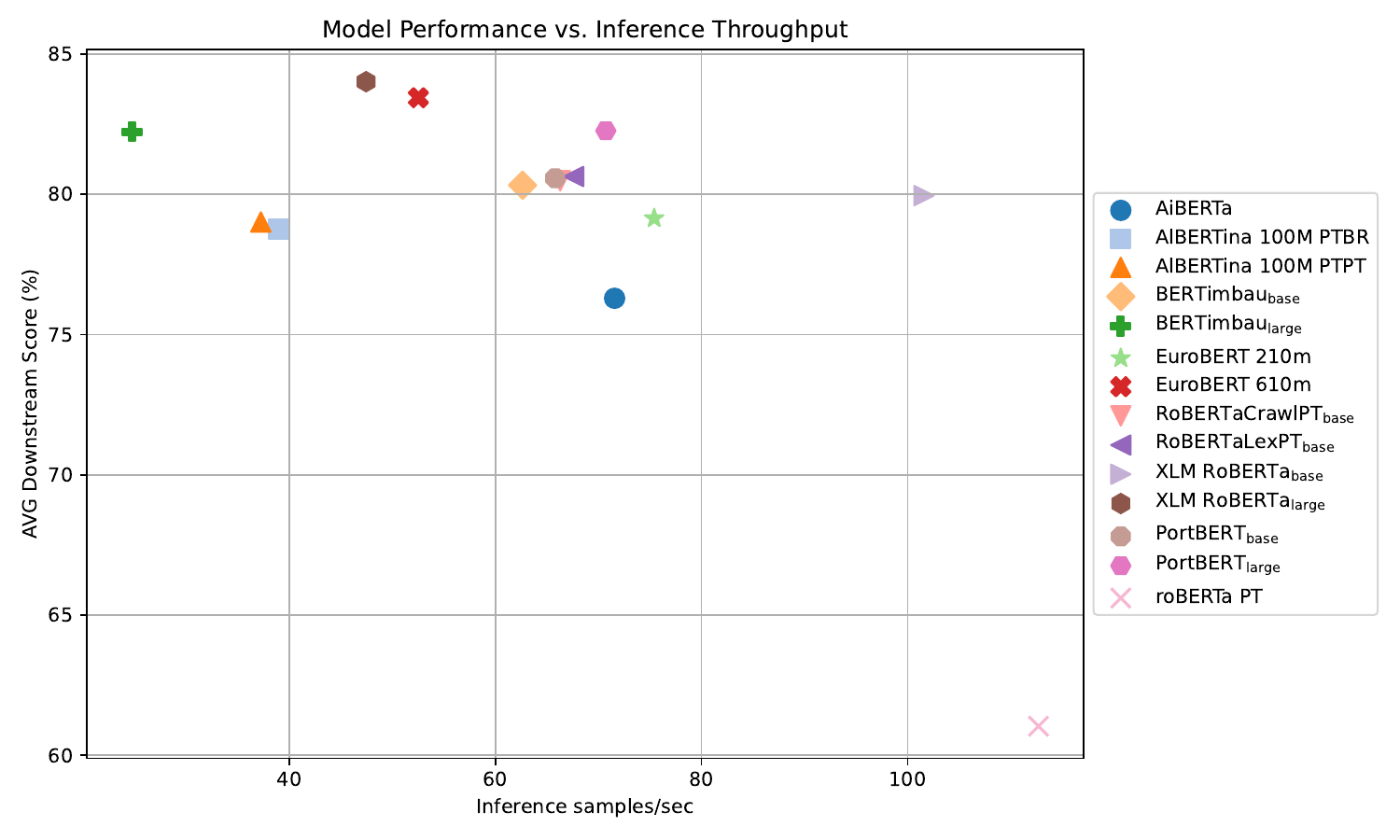}
  \caption {Performance–throughput trade-off across models. The top plot shows the relationship between average downstream score (AVG) and training throughput (samples/sec), while the bottom plot presents the same metric against inference throughput. This comparison highlights which models offer the best balance between effectiveness and computational efficiency during training and inference.}
    \label{fig:efficiency_plots}
\end{figure*}

To complement accuracy-based comparisons, we also assess model efficiency in terms of training and inference throughput (see Figure~\ref{fig:efficiency_plots}).
Among the base models, several exhibit a favorable balance between performance and efficiency. Notably, roBERTa PT achieves the highest training throughput (62.1 samples/sec) and inference speed (112.7 samples/sec), but its task performance lags significantly behind all competitors, suggesting that efficiency alone is insufficient without adequate pretraining quality. In contrast, PortBERT\textsubscript{base} and RoBERTaCrawlPT\textsubscript{base} both demonstrate strong downstream performance (AVG: 80.57 and 80.48, respectively) while maintaining competitive training throughput around 25–26 samples/sec and inference throughput above 65 samples/sec. BERTimbau\textsubscript{base} similarly offers a good trade-off with strong performance (AVG: 80.32) and respectable throughput, making these three the most efficient base models when balancing quality and compute.

The large models generally exhibit higher downstream performance but at a considerable computational cost. XLM RoBERTa\textsubscript{large} leads in task performance (AVG: 84.01) and inference throughput (47.4 samples/sec) compared to its large-model peers. However, its training throughput is relatively low (14.9 samples/sec), indicating longer training durations. PortBERT\textsubscript{large} achieves an attractive efficiency-performance trade-off, with an AVG of 82.26 while maintaining higher training and inference throughput (23.3 and 70.7 samples/sec, respectively), positioning it as the most throughput-efficient large model while still achieving competitive accuracy. Meanwhile, EuroBERT-610M delivers strong performance (AVG: 83.44) but with lower throughput metrics, reflecting its high computational demands. These results suggest that while large models provide superior accuracy, the efficiency gap between well-optimized base and large models like PortBERT is narrowing. Full runtime statistics are reported in Appendix~\ref{app:efficiency}.

\section{Discussion}
\subsection{Efficiency and Accuracy Trade-offs}
PortBERT demonstrates that efficient, monolingual transformer models remain a valuable asset in the evolving landscape of Portuguese NLP. While large multilingual encoders like XLM-RoBERTa or EuroBERT-610M offer strong performance, their high computational demands restrict practical deployment, particularly in latency-sensitive or resource-constrained settings. In contrast, PortBERT delivers competitive downstream task results while maintaining generally higher throughput compared to other strong Portuguese baselines, both during training and inference.

As shown in our efficiency analysis (Section~\ref{sec:efficiency}), PortBERT\textsubscript{base} stands out for its balanced trade-off between accuracy and efficiency, ranking among the top performers in its class. PortBERT\textsubscript{large} narrows the performance gap to state-of-the-art models like XLM RoBERTa\textsubscript{large}, while maintaining superior throughput and lower hardware demands.
% R2Q5: Is the difference of the developed models presented in this paper and XLM Roberta-large models due to just training data size?
Our focus with PortBERT was on cost-efficient pretraining for Portuguese specifically, where zero-shot transfer is not required. In this sense, PortBERT complements large multilingual encoders such as XLM-RoBERTa by offering a more efficient option for monolingual applications.

The performance differences between PortBERT and large multilingual encoders such as XLM-RoBERTa\textsubscript{large} are not solely attributable to the amount of training data. They also reflect architectural and training differences, including the substantially larger parameter count of XLM-RoBERTa (560M vs. 357M for PortBERT\textsubscript{large}), its much larger multilingual vocabulary (250k vs. 52k tokens), and the use of a massive multilingual corpus (2.5TB multilingual vs. 456GB of Portuguese).

In addition to hardware throughput, PortBERT models also demonstrate strong parameter efficiency. PortBERT\textsubscript{base} (126M parameters) achieves higher average performance than larger models such as XLM-RoBERTa\textsubscript{base} (278M) and EuroBERT-210M (212M), despite having less than half their parameter count. PortBERT\textsubscript{large} (357M) achieves results close to XLM-RoBERTa\textsubscript{large} (560M) and EuroBERT-610M (608M), highlighting the impact of targeted, monolingual pretraining on recent Portuguese corpora. This makes PortBERT a compelling choice in scenarios where both accuracy and model size matter.

\subsection{Training Setup and Hardware Comparisons}
Beyond per-job throughput, the total pretraining time differed substantially between hardware setups. PortBERT\textsubscript{base}, trained on 8 NVIDIA A40 GPUs, required approximately 27 days to complete 100k update steps. In contrast, PortBERT\textsubscript{large}, trained on a TPUv4 128 pod, completed training in just over 6 days. Both models used the same batch size, corpus, and optimizer settings in full precision (fp32), allowing for a clean comparison of training performance across hardware platforms. Using \GottBERTt's pretraining durations as a reference, we estimate that PortBERT\textsubscript{base} would have taken around 1.3 days to train on comparable TPU infrastructure. This illustrates the advantage of modern TPUs for large-scale training, particularly when time is a critical factor. However, TPU-specific constraints, including limited memory flexibility and less mature tooling for PyTorch and custom workflows, can limit development. In addition, the lack of local TPU hardware forces developers to rely on cloud platforms, slowing iteration and complicating debugging.

Efficiency comparisons must also consider hardware configuration. Due to memory constraints, EuroBERT-610M and partly XLM RoBERTa\textsubscript{large} were trained without parallel jobs (i.e., one job per GPU), whereas PortBERT and other models used multiple parallel training jobs per GPU to maximize utilization. This difference in hardware allocation might have impacted the observed throughput and training durations, potentially skewing efficiency comparisons in this regard.

\subsection{Positioning Among Existing Models}
Recent large-scale efforts such as EuroBERT~\citep{boizard2025eurobertscalingmultilingualencoders} illustrate the scale-performance frontier in multilingual modeling. EuroBERT training consumed over 200,000 GPU hours across MI250X and MI300A clusters and leveraged cutting-edge optimization techniques such as FlashAttention~\citep{dao2023flashattention2fasterattentionbetter}. While such models raise the performance ceiling, they also require infrastructure that is out of reach for many academic or industry teams. In contrast, PortBERT was trained on commodity hardware using open-source tools, offering a transparent and efficient alternative that lowers the entry barrier for building high-quality models in any languages.

To our knowledge, PortBERT is the first RoBERTa-style Portuguese model trained on recent deduplicated and filtered corpora from CulturaX (mC4) and OSCAR23, using a fully transparent and reproducible fairseq pipeline. This positions it as a strong alternative to more resource-intensive systems, particularly for researchers and practitioners seeking open, efficient solutions.

Although decoder-only models such as GPT variants dominate general-purpose NLP, they are often unsuitable for sentence-level classification tasks due to their autoregressive nature. Encoder-based models like PortBERT offer lower inference latency and better fit for downstream classification, especially under real-world constraints.

\subsection{Architectural Constraints and Training Stability}
% R1Q1: Why did you retain the original RoBERTa architecture without integrating newer architectural innovations (e.g., sparse attention, flash attention, etc.)?
We deliberately retained the standard RoBERTa encoder architecture. Our goal was not only to establish a strong monolingual baseline, but also to enable a fair comparison of computational costs with GottBERT, which was trained on a comparable TPU setup. Introducing architectural modifications such as sparse or FlashAttention would have shifted the baseline and made this comparison meaningless.

Like GeistBERT~\citep{scheibleschmitt2025geistbertbreathinglifegerman}, PortBERT prioritizes practical usability over raw scale. Although it does not achieve top performance on every benchmark, it remains consistently strong across tasks, making it a compelling option in the accuracy-efficiency trade-off. PortBERT could also be adapted for longer inputs using architectures such as Longformer~\citep{beltagy2020longformerlongdocumenttransformer} or Nyströmformer~\citep{xiong2021nystromformernystrombasedalgorithmapproximating}, though at the cost of increased training complexity.

% Did authors experiment with Whole Word Masking (WWM)? If not, why was it omitted despite its known benefits in token-level tasks?
During pretraining, we did not apply WWM, as stable support for it was missing in the fairseq TPU implementation. As with GottBERT, we encountered TPU-specific constraints: the lack of dynamic memory allocation required processing the corpus as a continuous token stream, deviating from RoBERTa’s dynamic sentence-sampling strategy. We were also constrained to 32-bit precision due to unstable 16-bit support in fairseq’s TPU implementation, increasing memory use and runtime. To ensure stability under these conditions, we used conservative learning rates. For comparability, we deliberately applied the same pre-processing and training constraints to the GPU-based base model, even though the GPU setup would have supported dynamic sampling and mixed precision.

\subsection{Final Remarks}
Ultimately, PortBERT is a step toward sustainable and accessible language modeling for Portuguese. It illustrates that thoughtful model design, combined with optimized pretraining and recent corpora, can yield strong models without relying on large-scale infrastructure. Future work may explore quantized or distilled versions for mobile deployment and domain-specific continued pretraining to further expand applicability or even continue pre-training with a more diverse corpus using WWM similar to \citet{scheibleschmitt2025geistbertbreathinglifegerman}.

\section{Conclusion}
We presented PortBERT, a family of RoBERTa-based language models for Portuguese, pre-trained on recent large-scale corpora (mC4 and OSCAR23). While not state-of-the-art on all benchmarks, PortBERT models achieve strong downstream performance and demonstrate notable efficiency in training and inference.
To support reproducibility and downstream adoption, we release both Huggingface-compatible models and fairseq checkpoints. These resources enable further pretraining, fine-tuning, or adaptation for longer contexts and domain-specific tasks. PortBERT offers an efficient and accessible foundation for Portuguese NLP.

\section*{Acknowledgments}
We gratefully acknowledge the support of Google’s TensorFlow Research Cloud (TFRC) for providing access to Cloud TPUs, which enabled efficient pretraining of PortBERT\textsubscript{large}. We also thank Nora Limbourg, our Google Cloud Customer Engineer, for her valuable technical assistance and coordination throughout the project.

R.S. would also like to thank Bruno \& Suzi, as well as all members and friends of the Best Spot Azores Diving Center, including Alberto \& Simona, Arturo, João \& Claudia, Maëlle \& Elias, Maria, Oliver, Paula, Raquel, Ruben, Sara and Vasco. Their kindness, presence, and community spirit provided strength and stability in a time of personal challenge. It is always a pleasure to dive with us.

\section*{Limitations}
This work has several limitations. First, although we used deduplicated and filtered corpora from CulturaX (mC4 and OSCAR23), we did not apply deduplication across all possible data sources or levels of granularity. Residual duplication or noise may therefore remain in the training data.

Second, PortBERT was trained exclusively on web-based Portuguese text, without explicit control for dialectal variation (e.g., Brazilian vs. European Portuguese) or domain-specific content. As a result, the model's performance on underrepresented dialects or specialized registers (e.g., legal, medical, or informal language) may be suboptimal without further fine-tuning.

Third, while we aimed for stable and reproducible training configurations across both GPU and TPU platforms, we opted for conservative learning rates and default precision settings to ensure stability, particularly on TPUs where dynamic memory allocation and mixed precision remain limited in fairseq. We did not explore extensive hyperparameter tuning in regard of the peak learning rate and did not apply WWM, which could potentially yield further gains.

% R2Q2: There should be a proper error analysis section showing where the model fails.
Fourth, we did not include a detailed error analysis of model predictions. While such an analysis could provide additional insights into systematic failure modes, our focus in this work was on efficiency and establishing strong baselines for Portuguese NLP.

Lastly, our evaluation is focused on the ExtraGLUE benchmark. While this provides a useful proxy for general NLP performance in Portuguese, it does not capture the full range of downstream tasks or real-world deployment settings. Moreover, ExtraGLUE does not offer a held-out test set with a submission server, which limits the ability to conduct blind evaluations and compare models in a standardized manner.

\section*{Ethical Considerations}
As with any large-scale language model, PortBERT is susceptible to inheriting and reproducing biases present in its training data. While we apply deduplication techniques to reduce noise and redundancy, deeper societal, cultural, and representational biases may persist. This is particularly relevant for downstream applications in sensitive domains such as healthcare, education, or public administration, where biased outputs could reinforce inequality or misinformation.

Training on large-scale web-based corpora also introduces privacy concerns. Although the dataset is filtered and preprocessed, models may inadvertently memorize and surface sensitive or personal information. Careful handling is necessary when deploying PortBERT in real-world applications, especially those involving user data or decision-making contexts.

Finally, despite efforts to balance performance and efficiency, pretraining transformer models on GPUs and TPUs consumes substantial computational resources. The associated energy usage and environmental impact underline the importance of developing sustainable training practices and promoting model reuse.

\bibliographystyle{acl_natbib}
\bibliography{anthology,ranlp2025}

\appendix

\section{Parameters}
The parameter space for our grid search is listed in Table \ref{tab:hyperparams}. In addition, Table \ref{tab:best_params} shows the parameters of the best models (selection based on validation set) of the respective tasks. We include these details to support reproducibility of our downstream results.

\begin{table}[htb]
    \centering
    \begin{tabular}{lc}
         \bfseries Parameter & \bfseries Values\\
         \hline
         Learning Rate & 7e-5, 5e-5, 2e-5, \\
                     & 1e-5, 7e-6, 5e-6, 1e-6 \\
         Batch Size & 16, 32, 48, 64 \\
         Epochs & 10 \\
         \hline
    \end{tabular}
    \caption{Hyperparameters used in the grid search of the downstream tasks.}
    \label{tab:hyperparams}
\end{table}

\begin{table*}[!htbp]
\label{tab:best_params_portbert}
\centering
\begin{tabular}{lcccccccc}%
    \hline
    \multirow{2}{*}{\bfseries Model} & 
    \multicolumn{2}{c}{\bfseries STS-B} & 
    \multicolumn{2}{c}{\bfseries RTE} & 
    \multicolumn{2}{c}{\bfseries WNLI} & 
    \multicolumn{2}{c}{\bfseries MRPC} \\
    \cmidrule{2-9}
     & BS & LR & BS & LR & BS & LR & BS & LR 
    \\
    \csvreader[late after line=\\]{hyperparams_portbert.csv}{}%
     {\csvcoli & \csvcolii & \csvcoliii & \csvcoliv & \csvcolv & \csvcolvi & \csvcolvii & \csvcolviii & \csvcolix}
    \hline
\end{tabular}
\caption{\label{tab:best_params}Hyperparameters of the best downstream task models for each task and pre-trained model. BS refers to batch size, and LR denotes the learning rate.}
\end{table*}

\section{Perplexity}
\label{sec:perplexity}

\begin{figure}[htb]
    \centering
    \begin{subfigure}[b]{0.5\textwidth}
         \centering
        \includegraphics[width=\columnwidth]{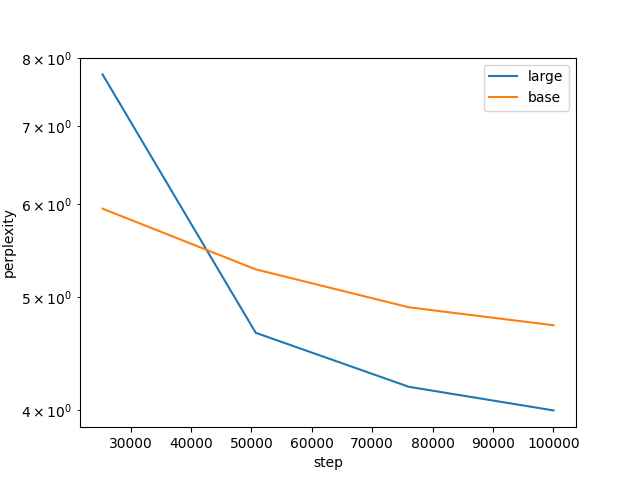}
    \end{subfigure}
    \begin{subfigure}[b]{0.5\textwidth}
         \centering
         \includegraphics[width=\columnwidth]{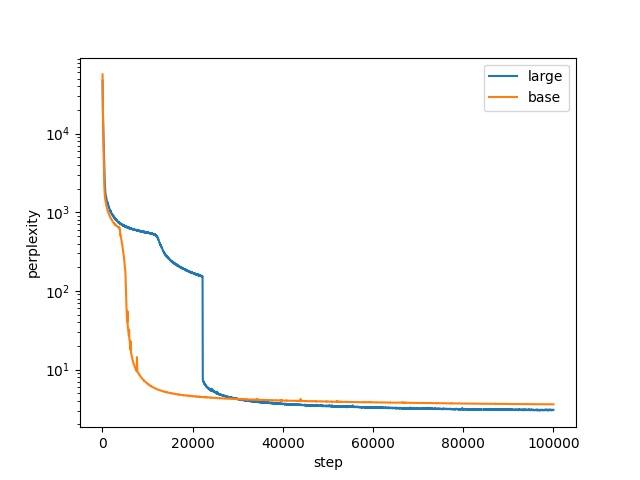}
    \end{subfigure}
    \caption{\label{fig:perplexity}Perplexity of the PortBERT models. Top based on a validation at the checkpoints. Bottom based on the validation of each optimization cycle during the training.}
\end{figure}

During pretraining, model perplexity was tracked on a test set after each optimization step and on a validation set at every checkpoint (see Figure~\ref{fig:perplexity}). The models exhibited a plateau in their perplexity curves, brief for the base models, but more prolonged for the large ones. Some training curves also showed temporary spikes, which may appear as divergence if not interpreted with context. Across both models, convergence occurred gradually and stabilized by around 30k steps. In contrast, the validation perplexity decreased steadily across both models without showing pronounced plateaus, stabilizing at lower values by the end of training. This results from the limited number of validation checkpoints (three intermediate epochs and a final checkpoint at 100k steps), which yield a coarser view of the learning dynamics.

\section{Efficiency Measurements}
\label{app:efficiency}

Tables~\ref{tab:gpu_time} and~\ref{tab:efficiency_table} report detailed runtime statistics for all models and tasks. Table~\ref{tab:gpu_time} provides a task-level breakdown of training and inference times, while Table~\ref{tab:efficiency_table} compares model-level efficiency metrics, including throughput and per-epoch timing. All models were fine-tuned using Huggingface Transformers (v4.52.3) on NVIDIA A40 GPUs.

\begin{table}[H]
    \centering
    \begin{tabular}{lcc}
        \hline
        \bfseries Task & \bfseries Training Time & \bfseries Inference Time \\
        \hline
        MRPC  & 157:04 & 00:38 \\
        RTE   & 241:46 & 00:57 \\
        STSB  & 314:24 & 02:25 \\
        WNLI  & 25:30  & 00:08 \\
        \hline
    \end{tabular}
    \caption{Computation time in hours and minutes for the downstream tasks, summing up to 1549 hours and 29 minutes, which corresponds to approximately 64.6 days of GPU usage.}
    \label{tab:gpu_time}
\end{table}

\begin{table*}[htb]
    \centering
    \begin{tabular}{lrrrrr}
        \toprule
        \textbf{Model} & \textbf{Train Time (s)} & \textbf{Train/s} & \textbf{Time/Epoch (s)} & \textbf{Eval Time (s)} & \textbf{Eval/s} \\
        \midrule
        AiBERTa\textsubscript{2000M}     & 1306.47 & 29.68 & 142.39 & 7.24  & 71.56 \\
        AlBERTina\textsubscript{PTBR}    & 2906.82 & 15.44 & 309.08 & 15.85 & 38.92 \\
        AlBERTina\textsubscript{PTPT}    & 2800.95 & 17.68 & 300.35 & 17.54 & 37.22 \\
        BERTimbau\textsubscript{base}    & 1499.94 & 25.12 & 152.88 & 9.15  & 62.63 \\
        BERTimbau\textsubscript{large}   & 4406.49 & 8.32  & 484.90 & 21.44 & 24.73 \\
        EuroBERT\textsubscript{210M}     & 1777.84 & 20.01 & 181.90 & 6.60  & 75.40 \\
        EuroBERT\textsubscript{610M}     & 2498.58 & 15.58 & 254.26 & 12.80 & 52.52 \\
        RoBERTaCrawlPT\textsubscript{base} & 1682.64 & 25.51 & 171.76 & 9.15  & 66.28 \\
        RoBERTaLexPT\textsubscript{base} & 1457.99 & 27.86 & 149.42 & 8.84  & 67.58 \\
        XLM-RoBERTa\textsubscript{base}  & 1440.55 & 24.97 & 152.49 & 4.86  & 101.59 \\
        XLM-RoBERTa\textsubscript{large} & 2139.34 & 14.85 & 233.65 & 10.46 & 47.44 \\
        PortBERT\textsubscript{base}     & 1524.59 & 25.00 & 160.29 & 8.96  & 65.79 \\
        PortBERT\textsubscript{large}    & 2389.63 & 23.26 & 264.74 & 15.09 & 70.71 \\
        roBERTa PT                       & 635.46  & 62.11 & 79.02  & 4.82  & 112.71 \\
        \bottomrule
    \end{tabular}
    \caption{Training and inference efficiency of all evaluated models. Metrics include total training time, training samples per second, average time per epoch, total evaluation time, and evaluation throughput.}
    \label{tab:efficiency_table}
\end{table*}

\end{document}